\def\year{2020}\relax
\documentclass[letterpaper]{article} 
\usepackage{aaai20}  
\usepackage{times}  
\usepackage{helvet} 
\usepackage{courier}  
\usepackage[hyphens]{url}  
\usepackage{graphicx} 
\usepackage{graphicx}  
\usepackage{amsmath}
\usepackage{amssymb}
\usepackage{pifont}
\newcommand{\cmark}{\ding{51}}%
\let\vec\mathbf
\DeclareMathOperator*{\argmax}{arg\,max}
\usepackage{multirow}
\usepackage{rotating}
\usepackage{xcolor}
\usepackage{enumitem}
\usepackage{bm}

\title{KnowIT VQA: Answering Knowledge-Based Questions about Videos}
\author{Noa Garcia,\textsuperscript{\rm 1} Mayu Otani,\textsuperscript{\rm 2} Chenhui Chu,\textsuperscript{\rm 1} Yuta Nakashima\textsuperscript{\rm 1}\\ 
\textsuperscript{\rm 1}Osaka University, Japan \quad
\textsuperscript{\rm 2}CyberAgent, Inc., Japan\\
\{noagarcia, chu, n-yuta\}@ids.osaka-u.ac.jp \\
otani\_mayu@cyberagent.co.jp 
}

\begin{document}

\maketitle

\begin{abstract}
We propose a novel video understanding task by fusing knowledge-based and video question answering. First, we introduce KnowIT VQA, a video dataset with 24,282 human-generated question-answer pairs about a popular sitcom. The dataset combines visual, textual and temporal coherence reasoning together with knowledge-based questions, which need of the experience obtained from the viewing of the series to be answered. Second, we propose a video understanding model by combining the visual and textual video content with specific knowledge about the show. Our main findings are: (i) the incorporation of knowledge produces outstanding improvements for VQA in video, and (ii) the performance on KnowIT VQA still lags well behind human accuracy, indicating its usefulness for studying current video modelling limitations.
\end{abstract}


\section{Introduction}
\label{sec:intro}

\begin{figure}
\centering
\includegraphics[width = 0.45\textwidth]{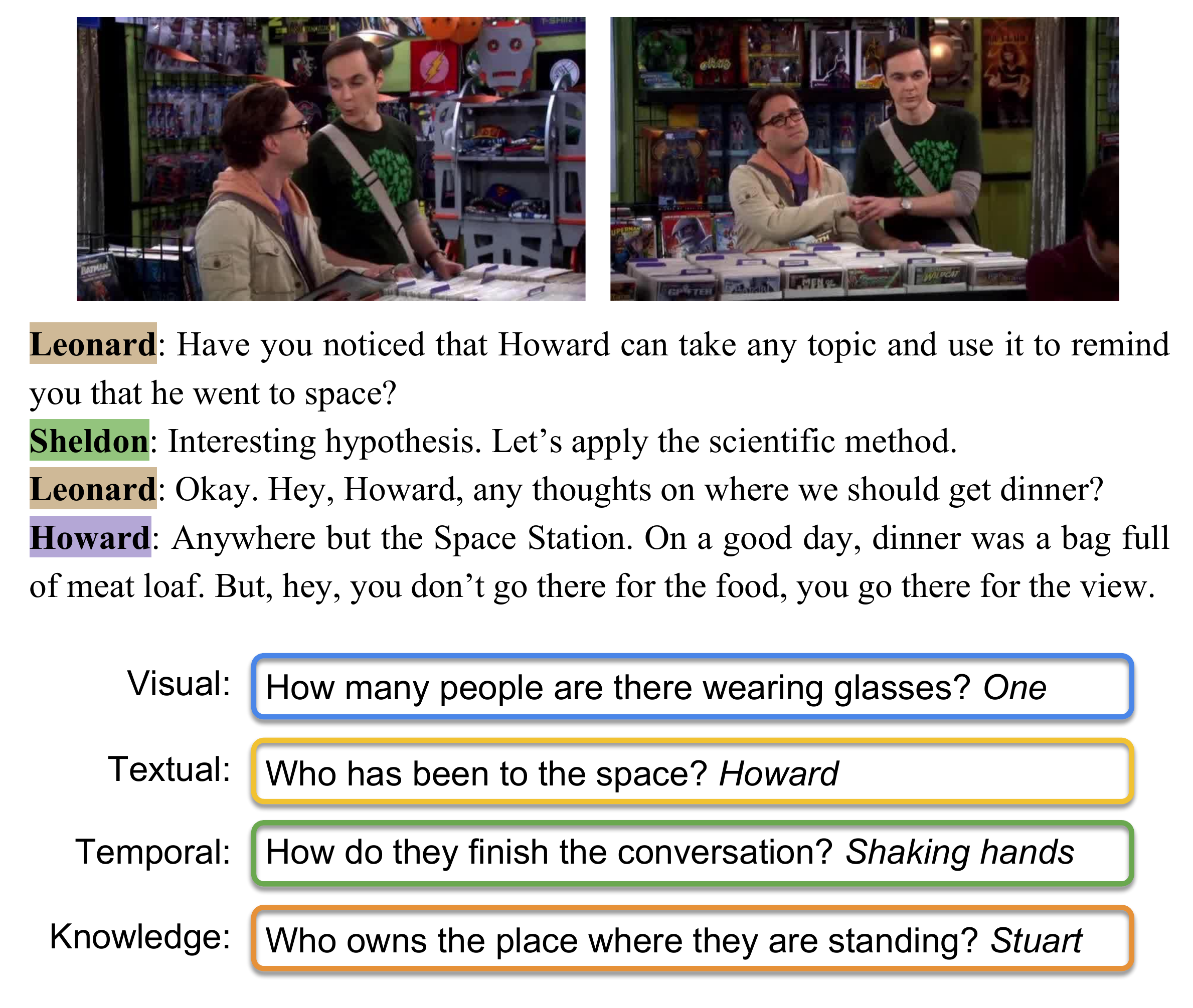}
\caption{Types of questions addressed in KnowIt VQA.} 
\label{fig:vqa_types}
\end{figure}

Visual question answering (VQA) was firstly introduced in \cite{malinowski2014multi} as a task for bringing together advancements in natural language processing and image understanding. Since then, VQA has experienced a huge development, in part due to the release of a large number of datasets, such as \cite{malinowski2014multi,antol2015vqa,krishna2017visual,johnson2017clevr,goyal2017making}. The current trend for addressing VQA \cite{Anderson2017up-down,kim2016hadamard,ben2017mutan,bai2018deep} is based on predicting the correct answer from a multi-modal representation, obtained from encoding images with a pre-trained convolutional neural network (CNN) and attention mechanisms \cite{xu2015show}, and encoding questions with a recurrent neural network (RNN). These kinds of models infer answers by focusing on the content of the images (e.g. \textit{How many people are there wearing glasses?} in Fig.~\ref{fig:vqa_types}). 

Considering that the space spanned by the training question-image pairs is finite, the use of image content as the only source of information to predict answers presents two important limitations. On one hand, image features only capture the static information of the picture, leaving temporal coherence in video unattended (e.g. \textit{How do they finish the conversation?} in Fig.~\ref{fig:vqa_types}), which is a strong constraint in real-world applications. On the other hand, visual content by itself does not provide enough insights for answering questions that require knowledge (e.g. \textit{Who owns the place were they are standing?} in Fig.~\ref{fig:vqa_types}). To address these limitations, video question answering (VideoQA) \cite{tapaswi2016movieqa,kim2017deepstory,lei2018tvqa} and knowledge-based visual question answering (KBVQA) \cite{wu2016ask,wang2018fvqa} have emerged independently by proposing specific datasets and models. However, a common framework for addressing multi-question types in VQA is still missing.

\begin{table*}
\caption{Comparison of VideoQA and KBVQA datasets. Answers are either multiple-choice ($\text{MC}_\text{N}$ with N being the number of choices) or single word. Last four columns refer to the type of questions available in each dataset.}
\centering
\small
\begin{tabular}{ l l l r r c c c c c}
\hline
\textbf{Dataset} & \textbf{VQA-Type} & \textbf{Domain} & \textbf{\# Imgs} & \textbf{\# QAs} & \textbf{Answers} & \textbf{Vis}. & \textbf{Text}. & \textbf{Temp}. & \textbf{Know}.\\
\hline
MovieQA {\scriptsize \cite{tapaswi2016movieqa}} & Video & Movie & 6,771 & 14,944 & $\text{MC}_5$ & \cmark & \cmark & \cmark & -\\
KB-VQA {\scriptsize \cite{wang2017explicit}} & KB & COCO & 700 & 2,402 & Word & \cmark & - & - & \cmark\\
PororoQA {\scriptsize \cite{kim2017deepstory}} & Video & Cartoon & 16,066 & 8,913 & $\text{MC}_5$ & \cmark & \cmark & \cmark & -\\
TVQA {\scriptsize \cite{lei2018tvqa}} & Video & TV show & 21,793 & 152,545 & $\text{MC}_5$ & \cmark & \cmark & \cmark & -\\
R-VQA {\scriptsize \cite{lu2018r}} & KB & \small Visual Genome & 60,473 & 198,889 & Word  &\cmark & - & - & - \\
FVQA {\scriptsize \cite{wang2018fvqa}}  & KB & \small COCO, ImgNet & 2,190 & 5,826 & Word & \cmark & - & - & \cmark \\
KVQA {\scriptsize \cite{shah2019kvqa}} & KB & Wikipedia & 24,602 & 183,007 & Word  &\cmark & - & - & \cmark \\
OK-VQA {\scriptsize \cite{marino2019ok}} & KB & COCO &  14,031 & 14,055 & Word & \cmark & - & - & \cmark\\
KnowIT VQA {\scriptsize (Ours)} & VideoKB & TV show & 12,087 & 24,282 & $\text{MC}_4$ & \cmark & \cmark & \cmark & \cmark\\
\hline
\end{tabular}
\label{tab:datasets}
\end{table*}

The contribution of this work lies in this line, by introducing a general framework in which both video understanding and knowledge-based reasoning are required to answer questions. We first argue that a popular sitcom, such as The Big Bang Theory,\footnote{\url{https://www.cbs.com/shows/big\_bang\_theory/}} is an ideal testbed for modelling knowledge-based questions about the world. With this idea, we created KnowIT VQA,\footnote{Available at \url{https://knowit-vqa.github.io/}} a dataset for KBVQA in videos in which real-world natural language questions are designed to be answerable only by people who is familiar with the show. We then cast the problem as a multi-choice challenge, and introduce a two-piece model that (i) acquires, processes, and maps specific knowledge into a continuous representation inferring the motivation behind each question, and (ii) fuses video and language content together with the acquired knowledge in a multi-modal fashion to predict the answer.


\section{Related Work}
\label{sec:relatedwork}

\paragraph{Video Question Answering} VideoQA addresses specific challenges with respect to the interpretation of temporal information in videos, including action recognition \cite{maharaj2017dataset,jang2017tgif,zellers2018vcr,mun2017marioqa}, story understanding \cite{tapaswi2016movieqa,kim2017deepstory}, or temporal coherence \cite{zhu2017uncovering}. Depending on the video source, the visual content of videos may also be associated with textual data, such as subtitles or scripts, which provide an extra level of context for its interpretation. Most of the proposed datasets so far are mainly focused on either the textual or the visual aspect of the video, without exploiting the combination of both modalities. In MovieQA \cite{tapaswi2016movieqa}, for example, questions are mainly plot-focused, whereas in other collections, questions are purely about the visual content, such as action recognition in MovieFIB \cite{maharaj2017dataset}, TGIF-QA  \cite{jang2017tgif}, and MarioVQA \cite{mun2017marioqa}, or temporal coherence in Video Context QA \cite{zhu2017uncovering}). Only few datasets, such as PororoQA \cite{kim2017deepstory} or TVQA \cite{lei2018tvqa}, present benchmarks for exploiting multiple sources of information, requiring models to jointly interpret multi-modal video representations. Even so, reasoning beyond the video content in these kinds of approaches is complicated, as only the knowledge acquired in the training samples is used to generate the answer.

\paragraph{Knowledge-Based Visual Question Answering} Answering questions about a visual query by only using its content constrains the output to be inferred within the space of knowledge contained in the training set. Considering that the amount of training data in any dataset is finite, the knowledge used to predict answers in standard visual question answering is rather limited. In order to answer questions beyond the image content, KBVQA proposes to inform VQA models with external knowledge. The way of acquiring and incorporating this knowledge, however, is still in early stages. For example, \cite{zhu2015building} creates a specific knowledge base with image-focused data for answering questions under a certain template, whereas more generic approaches \cite{wu2016ask} extract information from external knowledge bases, such as DBpedia \cite{auer2007dbpedia}, for improving VQA accuracy. As VQA datasets do not envisage questions with general information about the world, specific KBVQA datasets have been recently introduced, including KB-VQA \cite{wang2017explicit} with question-images pairs generated from templates, R-VQA \cite{lu2018r} with relational facts supporting each question, FVQA \cite{wang2018fvqa} with supporting facts extracted from generic knowledge bases, KVQA \cite{shah2019kvqa} for entity identification, or OK-VQA \cite{marino2019ok} with free-form questions without knowledge annotations. Most of these datasets impose hard constraints on their questions, such as being generated by templates (KB-VQA) or directly obtained from existing knowledge bases (FVQA), being OK-VQA the only one that requires handling unstructured knowledge to answer natural questions about images. Following this direction, we present a framework for answering general questions that may or may not be associated with a knowledge base by introducing a new VideoQA dataset, in which questions are freely proposed by qualified workers to study knowledge and temporal coherence together. To the best of our knowledge, this is the first work that explores external knowledge questions in a collection of videos.


\section{KnowIT VQA Dataset}
\label{sec:dataset}
Due to the natural structure of TV shows, in which characters, scenes, and general development of the story can be known in advance, TV data has been exploited for modelling real-world scenarios in video understanding tasks \cite{Nagrani17b,frermann2018whodunnit}. We also rely on this idea and argue that popular sitcoms provide an ideal testbed to encourage progress in knowledge-based visual question answering, due to their additional facilities to model knowledge and temporal coherence over time. In particular, we introduce the KnowIT VQA dataset, (standing for \underline{know}ledge \underline{i}nformed \underline{t}emporal VQA), a collection of videos from The Big Bang Theory annotated with knowledge-based questions and answers about the show.

\subsection{Video Collection}

Our dataset contains both visual and textual video data. Videos are collected from the first nine seasons of The Big Bang Theory TV show, with 207 episodes of about 20 minutes long each. For the textual data, we obtained the subtitles directly from the DVDs. Additionally, we downloaded episode transcripts from a specialised website.\footnote{\url{https://bigbangtrans.wordpress.com/}} Whereas subtitles are annotated with temporal information, transcripts associate dialog with characters. We align subtitles and transcripts with dynamic programming so that each subtitle is annotated to both its speaker and its timestamp. Transcripts also contain scene information, which is used to segment each episode into video scenes. Scenes are split uniformly into 20 seconds clips, obtaining 12,264 clips in total.

\subsection{QA Generation}
To generate real-world natural language questions and answers, we used Amazon Mechanical Turk (AMT)\footnote{\url{https://www.mturk.com}}. We required workers to have a high knowledge about The Big Bang Theory and instructed them to write knowledge-based questions about the show. Our aim was to generate questions answerable only by people familiar with the show, whereas difficult for new spectators. For each clip, we showed workers the video and subtitles, along with a link to the episode transcript and summaries of all the episodes for extra context. Workers were asked to annotate each clip with a question, its correct answer, and three wrong but relevant answers. The QA generation process was done in batches of one season at a time in two different rounds. During the second round, we showed the already collected data for each clip in order to 1) get feedback on the quality of the collected data and 2) obtain a diverse set of questions. The QA collection process took about 3 months.

\subsection{Knowledge Annotations}
We define knowledge as the information that is not contained in a given video clip. To approximate the knowledge the viewers acquire by watching the series, we annotated each QA pair with expert information:

\begin{figure}
\centering
\includegraphics[width = 0.42\textwidth]{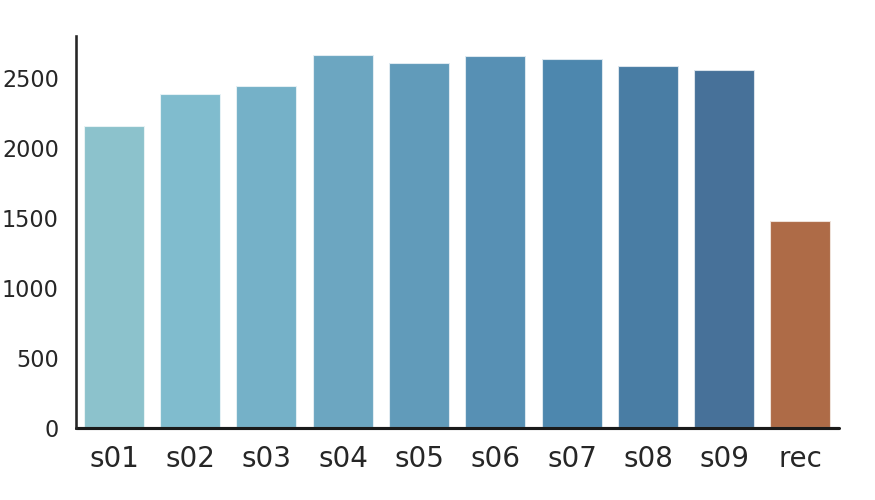}
\caption{Number of questions by \textsc{knowledge type}.} 
\label{fig:typereasdist}
\end{figure}

\begin{figure}
\centering
\includegraphics[width = 0.45\textwidth]{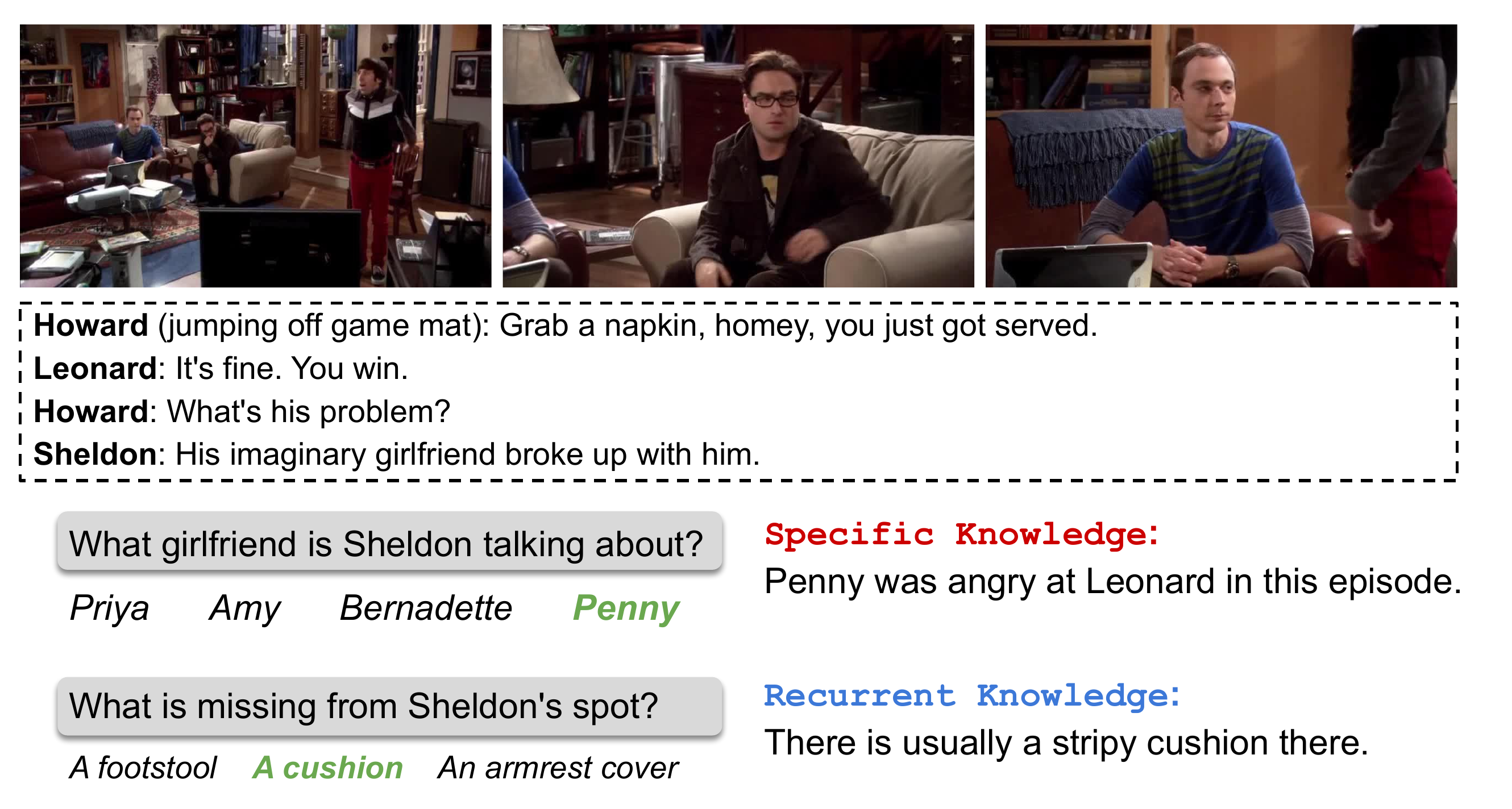}
\caption{Episode-specific versus recurrent \textsc{knowledge}.}
\label{fig:examplekgtype}
\end{figure}

\begin{figure}[t]
\centering
\includegraphics[width = 0.37\textwidth]{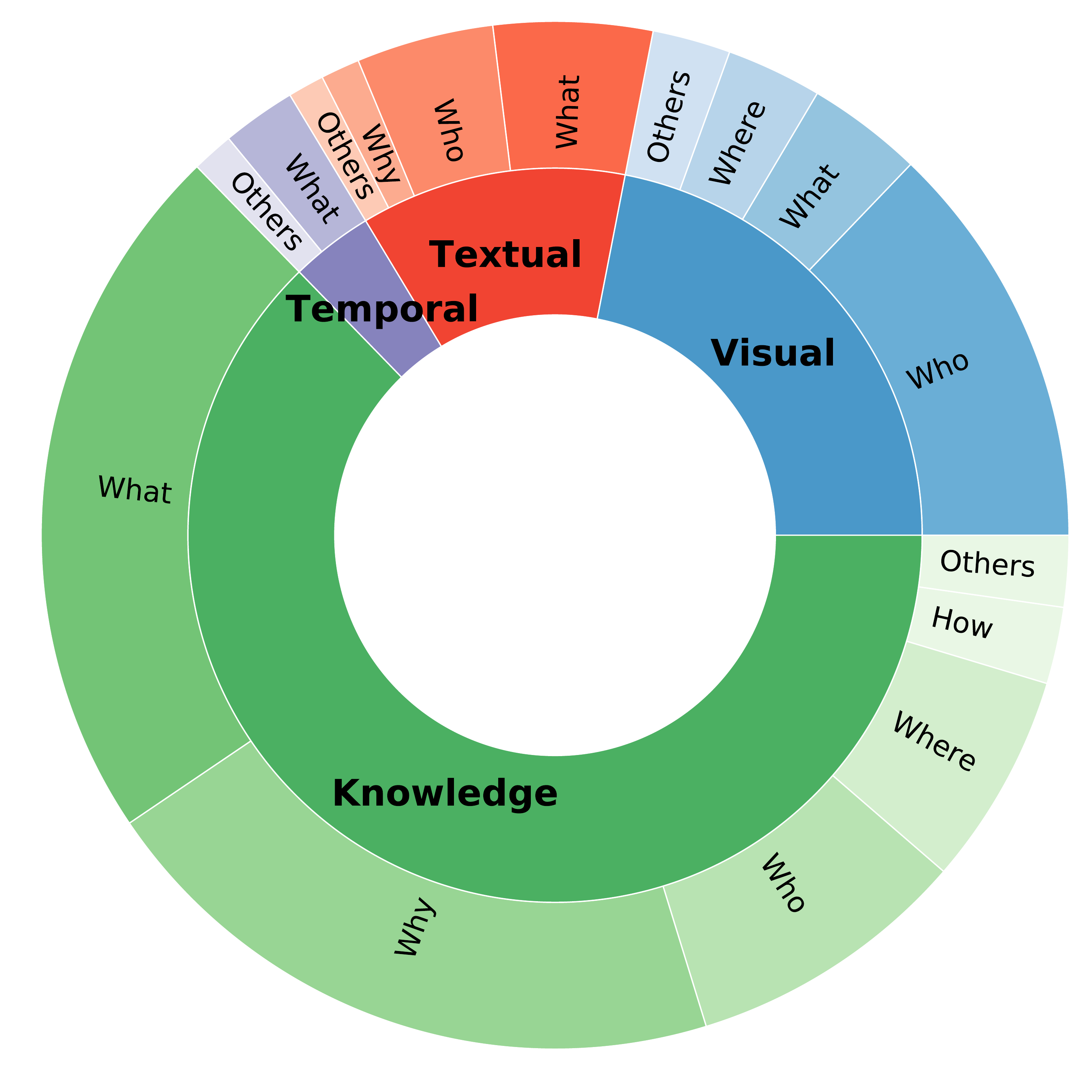}
\caption{Distribution of questions in the test set by their first word for each question type.}
\label{fig:qtypedist}
\end{figure}

\begin{itemize}
    \item \textsc{knowledge}: the information that is required to answer the question represented by a short sentence. For example, for the question \textit{Why did Leonard invite Penny to lunch?}, the information \textit{Penny has just moved in} is key to respond the correct answer, \textit{He wanted Penny to feel welcomed into the building}, over the other three  candidates.\footnote{1) \textit{Because he didn't have enough money to eat alone}, 2) \textit{Because he wanted Sheldon to practice his social skills}, and 3) \textit{Because he was in love with Penny}.}
    
    \item \textsc{knowledge type}: whether the knowledge is from the same episode (episode-specific) or it occurs repeatedly during the show (recurrent). The distribution between the two classes is shown in Fig.~\ref{fig:typereasdist}, with 6.08\% of the samples being recurrent and the rest being almost uniformly distributed over the nine seasons. Examples of recurrent and episode-specific \textsc{knowledge} are show in Figure \ref{fig:examplekgtype}.
    
    \item \textsc{question type}: we establish four different types of questions: 1) visual-based (22\%), in which the answer is found in the video frames, 2) textual-based (12\%), in which the answer is found in the subtitles, 3) temporal-based (4\%), in which the answer is predictable from the current video clip at a specific time, 4) knowledge-based (62\%), in which the answer is not found in the current clip, but in another sequence of the show. To encourage the development of general purpose models, \textsc{question type} is only provided for the test set. The distribution of question words in each type is plotted in Figure \ref{fig:qtypedist}.
\end{itemize}

\subsection{Data Splits}
We collected 24,282 samples from 12,087 video clips. We randomly split the episodes into training, validation, and test sets, so that questions and clips from the same episode were assigned to the same set. The number of episodes, clips, and QA pairs in each split are detailed in Table \ref{tab:datasplits}, as well as the average number of tokens in subtitles, questions, answers, and \textsc{knowledge}. Correct answers (CA) are slightly longer than wrong answers (WA), which is a common bias in QA datasets \cite{tapaswi2016movieqa,lei2018tvqa}.

\begin{table}
\caption{KnowIT VQA data splits and the average lengths.}
\centering
\begin{tabular}{ l r r r r}
\hline
& \textbf{Train} & \textbf{Val} & \textbf{Test} & \textbf{Total}\\
\hline
\# Episodes & 167 & 20 & 20 & 207 \\
\# Scenes & 2,007 & 225 & 240 & 2,472\\
\# Clips & 9,731 & 1,178 & 1,178 & 12,087\\
\# QAs & 19,569 & 2,352 & 2,361 & 24,282 \\
Len. Subtitles & 56.49 & 55.57 & 57.45 & 56.49 \\
Len. Questions & 7.5 & 7.38 & 7.48 & 7.49 \\
Len. CA & 4.55 & 4.51 & 4.46 & 4.54 \\
Len. WA & 4.14 & 4.12 & 4.06 & 4.13 \\
Len. \textsc{knowledge} & 10.43 & 10.10 & 10.30 & 10.39 \\
\hline
\end{tabular}
\label{tab:datasplits}
\end{table}

\subsection{Dataset Comparison} 
In Table \ref{tab:datasets}, we compare our dataset against other VideoQA and KBVQA datasets. KBVQA datasets are usually smaller than standard VQA datasets, as QA generation is often more challenging. Nevertheless, KnowIT VQA with 24k questions is the largest KBVQA human-generated dataset, far from the 2.4k questions in KB-VQA, 5.8k in FVQA, and 14k in OK-VQA. Also, KnowIT VQA is the  first collection addressing the four aforementioned types of questions. Note that the visual domain in KnowIT VQA is not new, sharing a small portion of videos (about 34\%) with TVQA. However, whereas TVQA uses 3.6k clips per show in average, the \textsc{knowledge} annotations in our dataset required a larger set of clips, in order to approximate the knowledge that spectators acquire by watching the series.


\section{Human Evaluation}
We performed human evaluation on the KnowIT VQA test set with a four-fold aim: 1) to evaluate whether video clips are relevant to answer questions; 2) to evaluate whether the questions \emph{do} require knowledge to be answered; 3) to evaluate whether the \textsc{knowledge} annotations are useful for answering the questions in the dataset; and 4) to introduce a human performance baseline for model comparison. 

\paragraph{Evaluation Design}
We used AMT with independent groups of workers for each task.\footnote{Workers participating on the creation of the dataset were not allowed to participate in the evaluation.} We split workers according to their experience with the show, i.e., \textit{masters}, who have watched at least the first nine seasons of the show, and \textit{rookies}, who have never watched any episode. We conducted two main tasks: evaluation on the questions and evaluation on the \textsc{knowledge} annotations.

\begin{figure}
\centering
\includegraphics[width = 0.45\textwidth]{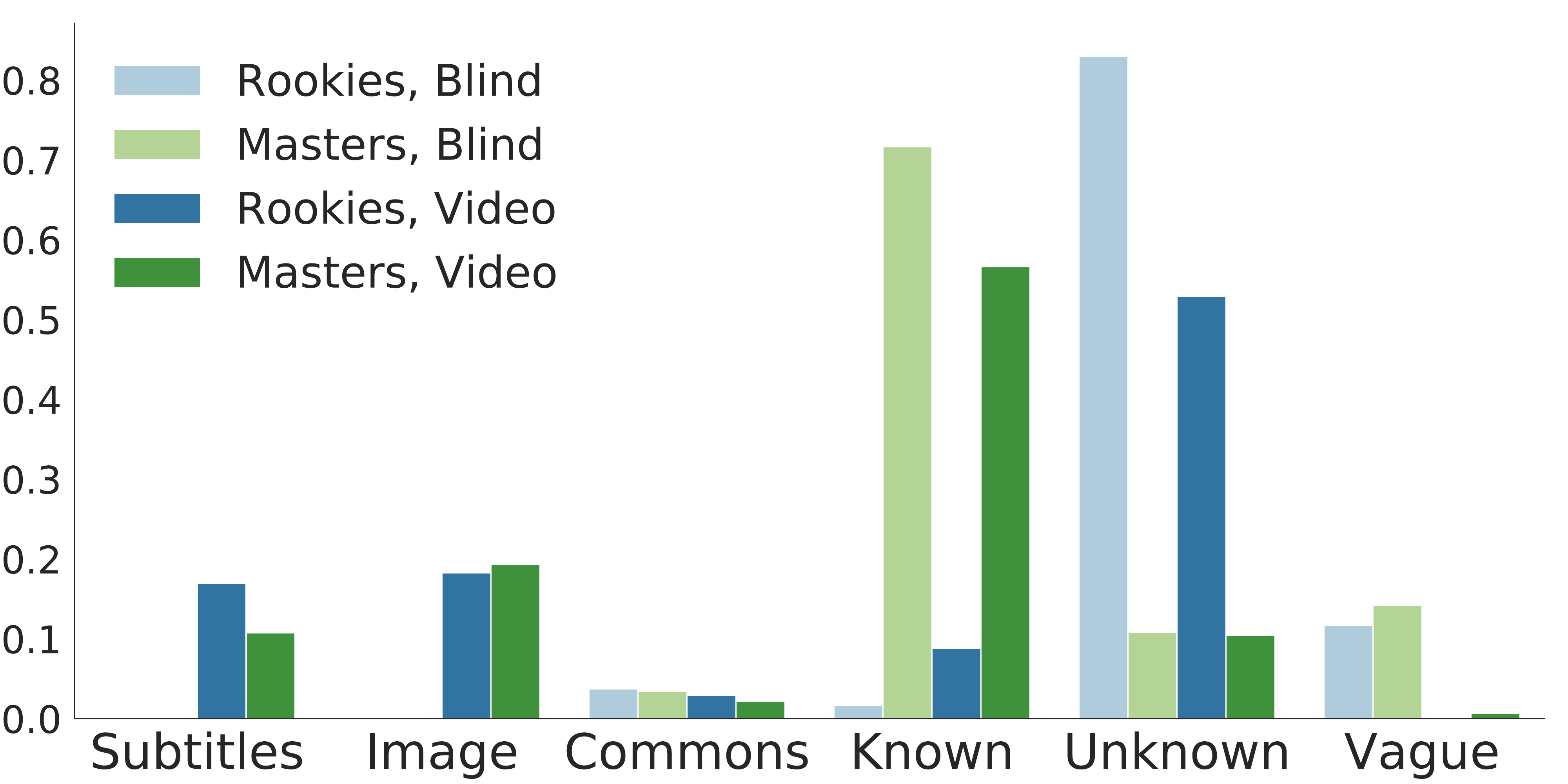}
\caption{Distribution of reasons for answering by groups.} \label{fig:human_reasons}
\end{figure}

\paragraph{Evaluation on the questions}
We further split masters and rookies into 3 different sub-groups according to the data provided to answer each question: Blind (only QAs), Subs (QAs and subtitles), and Video (QAs, subtitles, and clips). For each question in the test set, we asked workers to choose the correct answer from the four given candidates and to provide the reason for their response, from six possible options.\footnote{i) The answer is in the subtitles, ii) The answer is in the image, iii) The answer is common-sense knowledge, iv) I know the episode, v) I have no idea about the answer, and vi) The question is too vague to be answered.} In each group, each question was answered by 3 workers. Results are reported in Table \ref{tab:human}. The accuracy gap between Subs and Video groups confirms the relevance of the video content in the dataset. With respect to knowledge, the difference between masters and rookies strongly supports the claim that KnowIT VQA is extremely challenging when not knowing the show. When looking into the reasons for choosing the answer (Fig.~\ref{fig:human_reasons}), we saw that masters mostly based their choices on the knowledge acquired when watching the show, whereas rookies admitted not knowing the correct answer in most of their responses.

\begin{table}
\caption{Human evaluation on KnowIT VQA test set.}
\centering
\begin{tabular}{ l c c l c }
\hline
\textbf{Group} & \textbf{Acc} & & \textbf{Group} & \textbf{Acc}\\
\hline
Rookies, Blind & 0.440 & & Masters, Blind & 0.651 \\
Rookies, Subs & 0.562 & &  Masters, Subs & 0.789 \\
Rookies, Video & 0.748 & &  Masters, Video & 0.896 \\
\hline
\end{tabular}
\label{tab:human}
\end{table}

\begin{figure*}
\centering
\includegraphics[width = \textwidth]{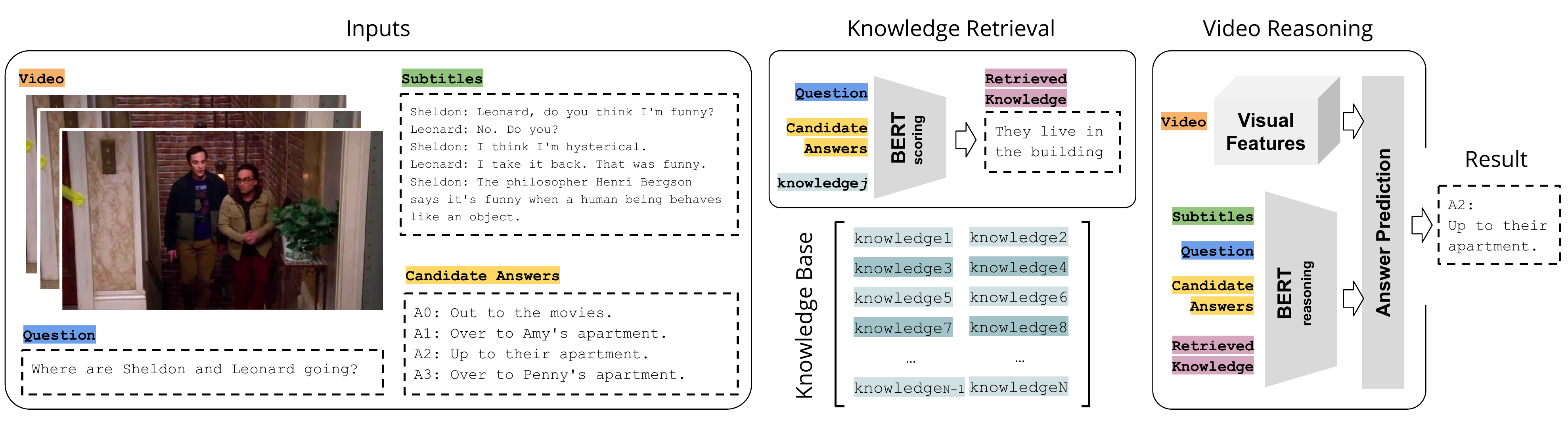}
\caption{Overview of \texttt{ROCK}. In the knowledge retrieval module, the question and candidate answers are used to retrieve knowledge instances in the KB with the BERT-scoring network. In the video reasoning part, visual features are extracted from the video, whereas subtitles, questions, candidate answers, and the retrieved knowledge instances are fed into the BERT-reasoning network. The visual and language representations are fused and fed into a classifier to predict the correct answer.} \label{fig:model}
\end{figure*}

\paragraph{Evaluation on the knowledge}
We studied the quality of the collected \textsc{knowledge} and its relevance to the questions in the dataset. We asked a group of rookies to answer the questions in the test set. For each question and candidate answers, we provided the subtitles and video clip. After answering, we showed the associated \textsc{knowledge} and asked them to answer again. As a result, we found that the accuracy increased from 0.683 (before \textsc{knowledge} exposition) to 0.823 (after \textsc{knowledge} exposition), verifying the relevance of the \textsc{knowledge} annotations to the questions.


\section{ROCK Model}
\label{sec:model}

We propose \texttt{ROCK} (Retrieval Over Collected Knowledge), a model for addressing knowledge-based visual question answering in videos, as depicted in Fig.~\ref{fig:model}. \texttt{ROCK} is based on the availability of language instances representing the show information in a knowledge base (KB), which \texttt{ROCK} retrieves and fuses with language and spatio-temporal video representations for reasoning over the question and predict the correct answer.

\subsection{Knowledge Base}
We first create a knowledge base (KB) to emulate the knowledge a viewer acquires when watching the series. Differently from previous work \cite{wu2016ask,wang2018fvqa}, which is based on generic knowledge graphs, such as DBpedia \cite{auer2007dbpedia}, our problem requires to access specific information about the show. Thus, we rely on the AMT workers annotations provided in the \textsc{knowledge} field during the dataset collection.\footnote{As future work, we plan to study how to automatically learn to generate similar explanations from another module that directly `watches' the series and extracts knowledge from videos.}

The collected \textsc{knowledge} is provided as natural language sentences. For example, to the question \textit{What was Raj doing at Penny's?}, the annotated \textsc{knowledge} is:

\vspace{7pt}

{\footnotesize \texttt{Raj wanted to ask Missy on a date, because Howard and Leonard had already asked her but failed, however his medication wore off and he couldn't do it.}}

\vspace{7pt}

As it is unclear how to capture such complex processes in an structured fashion such as knowledge graphs, we build a KB, $ K = \{w_j|j=1,\dots,N\}$, such that knowledge instances $w_j$'s  are represented as natural language sentences. We additionally perform a cleaning process to remove near-duplicate instances, reducing $N$ from 24,282 to 19,821. 

\paragraph{Knowledge Base Cleaning}
To remove near-duplicate samples in the KB, we compute similarities between \textsc{knowledge} instances. For each $w_j \in K$, we create an input sentence, $w'_{j}$, as a concatenation of strings: %
\begin{equation*}
  w'_{j} = \texttt{[CLS]} + w_j + \texttt{[SEP]},
\end{equation*}  %
where \texttt{[CLS]} is a token to indicate the start of the sequence and \texttt{[SEP]} is a token to indicate the end. We tokenise $ w'_{j}$ into a sequence of 60 tokens, $\boldsymbol{\tau}_{j}$. Let $\textsc{Bert}_\text{P}(\cdot)$ be a pre-trained BERT network \cite{devlin2018bert}, which takes as input a sequence of tokens and outputs the vector corresponding to the \texttt{[CLS]} token. We obtain the high-dimensional projection, $\vec{p}_{j}$, of $w_j$ as: %
\begin{equation}
 \vec{p}_{j} = \textsc{Bert}_\text{P}(\boldsymbol{\tau}_{j})
\end{equation}  %
To measure similarity between a pair of instances, ${w_i, w_j \in K}$, we compute a similarity score, $\beta_{ij}$, as: %
\begin{equation}
 \beta_{ij} = \text{sim}(\vec{p}_{i}, \vec{p}_{j}),
\end{equation}  %
where $\text{sim}(\cdot, \cdot)$ is the cosine similarity. Next, we build an undirected graph, ${G=(V,E)}$, in which nodes, ${V = \{w_j|j=1,\cdots,N\}}$, correspond to \textsc{knowledge} instances, and edges, $e = (w_i,w_j) \in E$, connect instances when $\beta_{ij} > 0.998$.\footnote{We experimentally found 0.998 to be a good tradeoff between near duplicates and semantically similar instances} To find near duplicate instances, we create clusters of nodes, $C_l$ with $l={1,\cdots,L}$, by finding all the connected components in $G$, i.e. $C_l$ corresponds to the $l$-th subgraph in $G$, for which all nodes are connected to each other by a path of edges. We randomly choose one node in each cluster and remove the others.


\subsection{Knowledge Retrieval Module}
Inspired by the ranking system in \cite{nogueira2019passage}, the knowledge retrieval module uses a question $q_i$ and its candidate answers $a^c_i$ with $c \in \{0,1,2,3\}$ to query the knowledge base $K$ and rank knowledge instances $w_j \in K$ according to a relevance score $s_{ij}$. 

We first obtain a sequence input representation $x_{ij}$ as a concatenation of strings: %
\begin{equation*}
   \resizebox{0.98\hsize}{!}{$x_{ij} = \texttt{[CLS]} + q_i + a^{\alpha_0}_i + a^{\alpha_1}_i + a^{\alpha_2}_i + a^{\alpha_3}_i + \texttt{[SEP]} + w_j + \texttt{[SEP]}$,}
\end{equation*}  %
where \texttt{[SEP]} separates the input text used for querying and the knowledge to be queried. Although preliminary experiments showed that the order of the answers $a^c_i$ does not have a high impact on the results, for an invariant model we automatically sort the answers according to a prior relevance score. $\alpha_{c}$ is then the original position of the answer with $c$-th highest score. Details are provided below. 

We tokenise $x_{ij}$ into a sequence of $n$ words $\vec{x}_{ij}$\footnote{Sequences longer than $n$ are truncated, and sequences shorter than $n$ are zero-padded.} and input it into a BERT network, namely BERT-scoring denoted by $\textsc{Bert}_\text{S}(\vec{x}_{ij})$, whose output is the vector corresponding to the \texttt{[CLS]} token. To compute $s_{ij}$, we use a fully connected layer together with a sigmoid activation as:%
\begin{equation}
s_{ij} = \text{sigmoid}(\vec{w}^\top_\text{S} \cdot \textsc{Bert}_\text{S}(\vec{x}_{ij}) + b_\text{S}),
\label{eq:scoreretrieval}
\end{equation} %
where $\vec{w}_\text{S}$ and $b_\text{S}$ are the weight vector and the bias scalar of the fully connected layer, respectively. $\textsc{Bert}^\text{S}_\theta$, $\vec{w}_\text{S}$, and $b_\text{S}$ are fine-tuned using matching (i.e. $i = j$) and non-matching (i.e. $i \neq j$) QA-knowledge pairs with the following loss: %
\begin{equation}
   \mathcal{L} =  - \sum_{i = j} \log(s_{ij}) - \sum_{i \neq j} \log(1 - s_{ij})
\end{equation} %

For each $q_i$, all $w_j$'s in $K$ are ranked according to $s_{ij}$. The top $k$ ranked instances, i.e. the most relevant samples for the query question, are retrieved.

\paragraph{Prior Score Computation}
To prevent the model producing different outputs for different candidate answer order, we create an answer order-invariant model by sorting answers, $a^c$ with $c=\{0,1,2,3\}$, according to a prior score, $\xi^c$.

For a given question $q$, $\xi^c$ is obtained from predicting the score of $a^c$ being the correct answer. We first build an input sentence $e^c$ as the concatenation of the strings: %
\begin{equation*}
  e^c = \texttt{[CLS]} + q + \texttt{[SEP]} + a^c + \texttt{[SEP]},
\end{equation*}  %
and we tokenise $e^c$ into a sequence of 120 tokens, $\vec{e}^c$. If $\textsc{Bert}_\text{E}(\cdot)$ represents a BERT network whose output is the vector corresponding to the \texttt{[CLS]} token, $\xi^c$ is obtained as: %
\begin{equation}
\xi^c = \vec{w}_\text{E}^\top \textsc{Bert}_\text{E}(\vec{e}^c) + b_\text{E},
\end{equation} %
Finally, all $\xi^c$ with $c=\{0,1,2,3\}$ are sorted in descending order into $\boldsymbol{\xi}$ and answers are ordered according to $\alpha_c = \delta$, where $\delta$ is the position of the $\delta$-th highest score in $\boldsymbol{\xi}$. 


\subsection{Video Reasoning Module}
In this module, the retrieved knowledge instances are jointly processed with the multi-modal representations from the video content to predict the correct answer. This process contains three components: visual representation, language representation, and answer prediction.

\paragraph{Visual Representation} 
We sample $n_f$ frames from each video clip and apply four different techniques to describe their visual content:

\begin{itemize}
\item \textit{Image features}: Each frame is fed into Resnet50 \cite{he2016deep} without the last fully-connected layer and is represented by a 2,048-dimensional vector. We concatenate all vectors from the $n_f$ frames and condense it into a 512-dimensional vector using a fully-connected layer.

\item \textit{Concepts features}: For a given frame, we use the bottom-up object detector \cite{Anderson2017up-down} to obtain a list of objects and attributes. We encode all the objects and attributes in the $n_f$ frames into a $C$-dimensional bag-of-concept representation, which is projected into a 512-dimensional space with a fully-connected layer. $C$ is the total number of available objects and attributes.

\item \textit{Facial features}: We use between 3 to 18 photos of the main cast of the show to train the state-of-the-art face recognition network in \cite{parkhi2015deep}.\footnote{Characters trained in the face recognition network are: Amy, Barry, Bernadette, Dr. Beverly Hofstadter, Dr.~VM Koothrappali, Emily, Howard, Leonard, Leslie, Lucy, Mary Cooper, Penny, Priya, Raj, Sheldon, Stuart, and Wil Wheaton.} For each clip, we encode the detected faces as a $F$-dimensional bag-of-faces representation, which is projected into a 512-dimensional space with a fully-connected layer. $F$ is the total number of people trained in the network.
 
\item \textit{Caption features}: For each frame, we generate a caption to describe its visual content using \cite{xu2015show}. The $n_f$ captions extracted from each clip are passed to the language representation model.
\end{itemize}

\paragraph{Language Representation} Text data is processed using a fine-tuned BERT model, namely BERT-reasoning. We compute the language input, $y^{c}$, as a concatenation of strings: %
\begin{equation*}
    \resizebox{0.98\hsize}{!}{$y^{c} = \texttt{[CLS]} + caps + subs + q + \texttt{[SEP]} + a^c + w + \texttt{[SEP]}$,}
\end{equation*} %
where $caps$ is the concatenated $n_f$ captions (ordered by timestamp), $subs$ the subtitles, and $w$ the concatenated $k$ retrieved knowledge instances. For each question $q$, four different $y^c$ are generated, one for each of the candidate answers $a^c$ with $c = \{0,1,2,3\}$. We tokenise $y^c$ into a sequence of $m$ words, $\vec{y}^c$, as in BERT-scoring. Let $\textsc{Bert}_\text{R}$ denote  BERT-reasoning, whose output is the vector corresponding to the \texttt{[CLS]} token. For $a^c$, the language representation $\vec{u}^c$ is obtained as $\vec{u}^c = \textsc{Bert}_\text{R}(\vec{y}^c)$.

\paragraph{Answer Prediction} To predict the correct answer, we concatenate the visual representation $\vec{v}$ (i.e.~image, concepts, or facial features) with one of the language representations $\vec{u}^c$: %
\begin{equation}
\vec{z}^c = [\vec{v}, \vec{u}^c],
\end{equation} %
$\vec{z}^c$ is projected into a single score, $o^c$, with a fully-connected layer: %
\begin{equation}
o^c = \vec{w}_\text{R}^\top \vec{z}^c + b_\text{R},
\end{equation} %
The predicted answer $\hat{a}$ is obtained with the index of the maximum value in ${\vec{o} = (o^0, o^1, o^2, o^3)^\top}$, i.e., ${\hat{a} = a^{\argmax_{c} \vec{o}}}$. Being $c^*$ the correct class, $\textsc{Bert}_\text{R}$, $\vec{w}_\text{R}$, and $b_\text{R}$ are fine-tuned with the multi-class cross-entropy loss as:%
\begin{equation}
\mathcal{L}(\vec{o}, c^*) =  - \log \frac{\exp(o^{c^*})}{\sum_c \exp(o^c)} 
\label{eq:crossentropyloss}
\end{equation} %


\section{Experimental Results}
We evaluated and compared \texttt{ROCK} against several baselines on the KnowIT VQA dataset. Results per question type and overall accuracy are reported in Table \ref{tab:results}. Models were trained with stochastic gradient descent with momentum of 0.9 and learning rate of 0.001. In BERT implementations, we used the uncased base model with pre-trained initialisation.

\paragraph{Answers} To detect potential biases in the dataset, we evaluated the accuracy of predicting the correct answer by only considering the candidate answers: 
\begin{itemize}
\item \texttt{Longest/Shortest}: The predicted answer is the one with the largest/smallest number of words.
\item \texttt{word2vec/BERT sim}: For word2vec, we use 300-dimensional pre-trained word2vec vectors \cite{mikolov2013distributed}. For BERT, we encode words with the output of the third-to-last layer of pre-trained BERT. Answers are encoded as the mean of their word representations. The prediction is the answer with the highest cosine similarity to the other candidates in average.
\end{itemize}
In general, these baselines performed very poorly, with only \texttt{Longest} being better than random. Other than the tendency of correct answers to be longer, results do not show any strong biases in terms of answer similarities.

\begin{figure*}
\centering
\includegraphics[width =\textwidth]{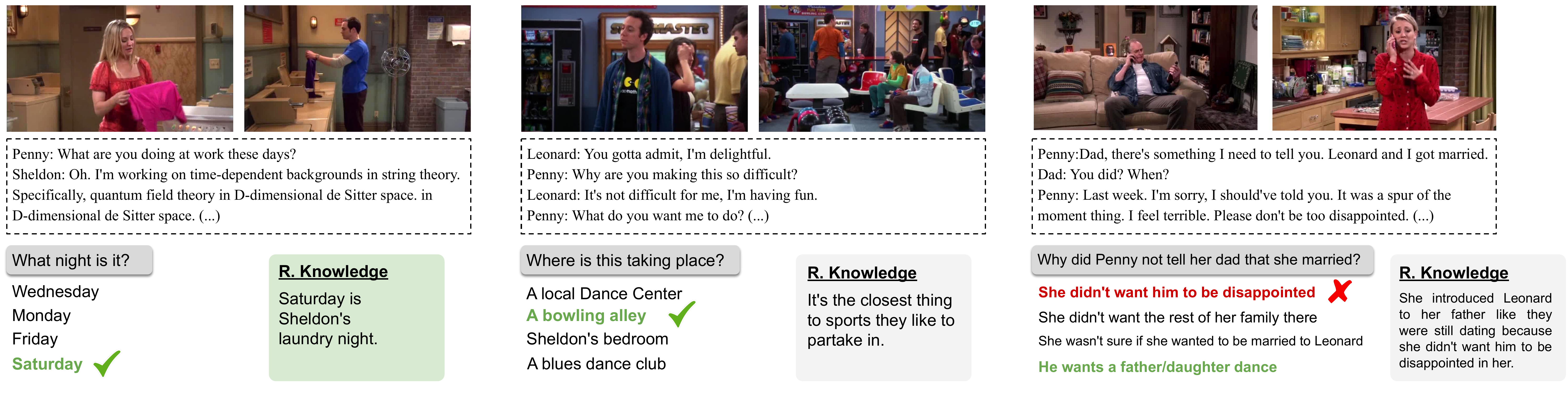}
\caption{Qualitative results of the \texttt{ROCK} {\footnotesize (Image)} model.  Left: the retrieved knowledge (RK) helps to predict the answer. Middle: the RK is not accurate, but the model still predicts the correct answer. Right: the RK is incorrect and leads to a misprediction.}
\label{fig:visualresults}
\end{figure*}

\paragraph{QA} We also evaluated several baselines in which only questions and candidate answers are considered. 
\begin{itemize}
\item \texttt{word2vec/BERT sim}: Questions and answers are represented by the mean word2vec or pre-trained BERT word representation. The predicted answer is the one with highest cosine similarity to the question. 
\item \texttt{TFIDF}: Questions and answers are represented as a weighted frequency word vector (tf-idf) and projected into a 512-dimensional space. The question and the four answer candidates are then concatenated and input into a four-class classifier to predict the correct answer.
\item \texttt{LSTM Emb./BERT}: Each word in a question or in a candidate answer is encoded through an embedding layer or a pre-trained BERT network and input into an LSTM \cite{hochreiter1997long}. The last hidden state of the LSTM is used as a 512-dimensional sentence representation. Question and answers are concatenated and input into a four-class classifier for prediction.
\item \texttt{ROCK\textsubscript{QA}}: \texttt{ROCK} model with $m=120$ tokens, trained and evaluated only with questions and answers as input.
\end{itemize}
Whereas methods based on sentence similarity performed worse than random, methods with classification layers trained for answer prediction (i.e.~\texttt{TFIDF}, \texttt{LSTM Emb./BERT}, and \texttt{ROCK\textsubscript{QA}}) obtained considerably better accuracy, even outperforming human workers.

\begin{table}
\caption{Accuracy for different methods on KnowIt VQA dataset. $\diamondsuit$ for parts of our model, $\bigstar$ for our full model.}
\centering
\resizebox{0.48\textwidth}{!}{\begin{tabular}{c l c c c c c }
\hline

& \textbf{Model} & \textbf{Vis.} & \textbf{Text.} & \textbf{Temp.} & \textbf{Know.} & \textbf{All}\\
\hline
& Random                      & 0.250 & 0.250 & 0.250 & 0.250 & 0.250 \\ 
\hline

\parbox[t]{3pt}{\multirow{4}{*}{\rotatebox[origin=c]{90}{Answers}}} 
 & \texttt{Longest}              & 0.324 & 0.308 & 0.395 & 0.342 & 0.336\\
 & \texttt{Shortest}             & 0.241 & 0.236 & 0.233 & 0.297 & 0.275\\
 & \texttt{word2vec sim}                & 0.166 & 0.196 & 0.233 & 0.189 & 0.186\\
 & \texttt{BERT sim}                    & 0.199 & 0.239 & 0.198 & 0.226 & 0.220\\
\hline

\parbox[t]{3pt}{\multirow{7}{*}{\rotatebox[origin=c]{90}{QA}}}
& \texttt{word2vec sim}                & 0.108 & 0.163 & 0.151 & 0.180 & 0.161 \\
& \texttt{BERT sim}                    & 0.174 & 0.264 & 0.209 & 0.190 & 0.196 \\
& \texttt{TFIDF}                       & 0.434 & 0.377 & 0.488 & 0.485 & 0.461 \\
& \texttt{LSTM Emb.}                   & 0.444 & 0.428 & 0.512 & 0.515 & 0.489 \\
& \texttt{LSTM BERT}                   & 0.446 & 0.464 & 0.500 & 0.532 & 0.504 \\
& $\diamondsuit$ \texttt{ROCK\textsubscript{QA}}   & 0.542 & 0.475 & 0.547 & 0.535 & 0.530 \\
& Humans {\scriptsize (Rookies, Blind)}            & 0.406 & 0.407 & 0.418 & 0.461 & 0.440 \\
\hline

\parbox[t]{3pt}{\multirow{5}{*}{\rotatebox[origin=c]{90}{Subs, QA}}}
& \texttt{LSTM Emb.}                   & 0.432 & 0.362 & 0.512 & 0.496 & 0.467 \\
& \texttt{LSTM BERT}                   & 0.452 & 0.446 & 0.547 & 0.530 & 0.504 \\
& \texttt{TVQA\textsubscript{SQA}}            & 0.602 & 0.551 & 0.512 & 0.468 & 0.509 \\
& $\diamondsuit$ \texttt{ROCK\textsubscript{SQA}}             & 0.651 & 0.754 & 0.593 & 0.534 & 0.587 \\
& Humans {\scriptsize (Rookies, Subs)}            & 0.618 & 0.837 & 0.453 & 0.498 & 0.562 \\
\hline

\parbox[t]{3pt}{\multirow{6}{*}{\rotatebox[origin=c]{90}{Vis, Subs, QA}}}
& \texttt{TVQA}     & 0.612 & 0.645 & 0.547 & 0.466 & 0.522 \\
& $\diamondsuit$ \texttt{ROCK\textsubscript{VSQA}} {\scriptsize Image}              & 0.643 & 0.739 & 0.581 & 0.539 & 0.587 \\
& $\diamondsuit$ \texttt{ROCK\textsubscript{VSQA}} {\scriptsize Concepts}              & 0.647 & 0.743 & 0.581 & 0.538 & 0.587 \\
& $\diamondsuit$ \texttt{ROCK\textsubscript{VSQA}} {\scriptsize Facial}              & 0.649 & 0.743 & 0.581 & 0.537 & 0.587 \\
& $\diamondsuit$ \texttt{ROCK\textsubscript{VSQA}} {\scriptsize Caption}              & 0.666 & 0.772 & 0.581 & 0.514 &  0.580\\
& Humans {\scriptsize (Rookies, Video)}           & 0.936 & 0.932 & 0.624 & 0.655 & 0.748 \\
\hline

\parbox[t]{3pt}{\multirow{6}{*}{\rotatebox[origin=c]{90}{Knowledge}}}
&  $\bigstar$ \texttt{ROCK} {\scriptsize Image}  & 0.654 & 0.681 & 0.628 & 0.647 & 0.652 \\
& $\bigstar$ \texttt{ROCK} {\scriptsize Concepts} & 0.654 & 0.685 & 0.628 & 0.646 & 0.652 \\
& $\bigstar$ \texttt{ROCK} {\scriptsize Facial} & 0.654 & 0.688 & 0.628 & 0.646 & 0.652 \\
& $\bigstar$ \texttt{ROCK} {\scriptsize Caption} & 0.647 & 0.678 & 0.593 & 0.643 & 0.646 \\
& \texttt{ROCK\textsubscript{GT}} & 0.747 & 0.819 & 0.756 & 0.708 &  0.731\\
& Humans {\scriptsize (Masters, Video)}             & 0.961 & 0.936 & 0.857 & 0.867 & 0.896 \\
\hline
\end{tabular}}
\label{tab:results}
\end{table}

\paragraph{Subs, QA} Models that use subtitles, questions, and answers as input. 
\begin{itemize}
    \item \texttt{LSTM Emb./BERT}: Subtitles are encoded with another LSTM  and concatenated to the question and answer candidates before being fed into the four-class classifier.
    \item \texttt{TVQA\textsubscript{SQA}} \cite{lei2018tvqa}: Language is encoded with a LSTM layer and no visual information is used.
    \item \texttt{ROCK\textsubscript{SQA}}: With $m = 120$ tokens, the input sequence only includes subtitles, questions, and candidate answers.
\end{itemize}
\texttt{LSTM BERT} and \texttt{ROCK\textsubscript{SQA}} improved accuracy by a 5.7\% with respect to only questions and answers. On the other hand, \texttt{LSTM Emb.} did not improve compared to the models using only QA, which may imply a limitation in the word embeddings to encode long sequences in subtitles.

\paragraph{Vis, Sub, QA} VideoQA models based on both language and visual representations. 
\begin{itemize}
    \item \texttt{TVQA} \cite{lei2018tvqa}: State-of-the-art VideoQA method. Language is encoded with a LSTM layer, whereas visual data is encoded into visual concepts.
    \item \texttt{ROCK\textsubscript{VSQA}}: Our model with $m = 120$ tokens and $n_f = 5$ frames. Four different visual representations are used.
\end{itemize}
\texttt{ROCK\textsubscript{VSQA}} outperformed \texttt{TVQA} by 6.6\%, being Concepts the features with the highest accuracy. However, any of the visual models outperformed \texttt{ROCK\textsubscript{SQA}}, implying strong limitations in current video modelling approaches.

\paragraph{Knowledge} Models that exploit \textsc{knowledge} to predict the correct answer, i.e.~our \texttt{ROCK} model in its full version, with $n=128$ and $k=5$ in the knowledge retrieval module, and $m=512$ in the video reasoning module. Compared to the non-knowledge methods, the inclusion of the knowledge retrieval module increased the accuracy by 6.5\%, showing the great potential of knowledge-based approaches in our dataset. Among the visual representations, Image, Concepts, and Facial performed the same. However, when compared against human masters, \texttt{ROCK} lags well behind, suggesting potential room for improvement. When using the annotated \textsc{knowledge} instead of the retrieved one (\texttt{ROCK}\textsubscript{GT}), accuracy is boosted to 0.731, indicating that improvements in the knowledge retrieval module will increase the overall performance. Finally, qualitative results are presented in Fig.~\ref{fig:visualresults}, providing some insights on the strengths and weaknesses of our model.

\paragraph{Knowledge Retrieval Results}
Results for the knowledge retrieval module, in terms of recall at K (R@K) and median rank (MR), are shown in Table \ref{tab:retrieval}. We tested different arrangements in the input data: 
\begin{itemize}
    \item Only Questions: Candidate answers were not used.
    \item QA parameter sharing: In the input string, $x_{ij}$, only one answer, $a^c$, at a time was used as $x_{ij} = \texttt{[CLS]} + q_i + a^c_i + \texttt{[SEP]} + w_j + \texttt{[SEP]}$, which means that the same parameters are used for the four candidate answers.
    \item QA prior score: Our proposed method based on ordering answers according to their prior score.
\end{itemize}
There was a big gap between Only Questions and the other two methods, indicating that candidate answers contained relevant information to retrieve the correct knowledge. The best results were obtained with our proposed prior scoring method, which showed that using all the candidate answers together provided more context for finding the correct knowledge instance.

\begin{table}
\caption{Knowledge retrieval module results on test set.}
\centering
\resizebox{0.48\textwidth}{!}{\begin{tabular}{ l c c c c c }
\hline
\textbf{Method} & \textbf{R@1} & \textbf{R@5} & \textbf{R@10} & \textbf{R@100} & \textbf{MR}\\
\hline
Only Questions & 0.070 & 0.169 & 0.208 & 0.426 & 221 \\
QA param sharing & 0.083 & 0.197 & 0.268 & 0.557 & 67 \\
QA prior score & 0.114 & 0.259 & 0.318 & 0.576 & 53 \\
\hline
\end{tabular}}
\label{tab:retrieval}
\end{table}


\section{Conclusion}
\label{sec:conclusion}

We presented a novel dataset for knowledge-based visual question answering in videos and proposed a video reasoning model, in which multi-modal video information was combined together with specific knowledge about the task. Our evaluation showed the great potential of knowledge-based models in video understanding problems. However, there is still a big gap with respect to human performance, which we hope our dataset will contribute to reduce by encouraging the development of stronger models.


\section{Acknowledgements}

This work is based on results obtained from a project commissioned by the New Energy and Industrial Technology Development Organization (NEDO). It was also partly supported by JSPS KAKENHI No.~18H03264 and  JST ACT-I. 

\bibliography{AAAI-GarciaN.1192}
\bibliographystyle{aaai}


\end{document}